*Automated Transcription of Non-Latin Script Periodicals: A Case Study in the Ottoman Turkish Print Archive*

Suphan Kirmizialtin, NYUAD
David Joseph Wrisley, NYUAD

# Abstract

Our study utilizes deep learning methods for the automated transcription of late nineteenth- and early twentieth-century periodicals written in Arabic script Ottoman Turkish (OT) using the *Transkribus* platform. We discuss the historical situation of OT text collections and how they were excluded for the most part from the late twentieth century corpora digitization that took place in many Latin script languages. This exclusion has two basic reasons: the technical challenges of OCR for Arabic script languages, and the rapid abandonment of that very script in the Turkish historical context. In the specific case of OT, opening periodical collections to digital tools require training HTR models to generate transcriptions in the Latin writing system of contemporary readers of Turkish, and not, as some may expect, in right-to-left Arabic script text. In the paper we discuss the challenges of training such models where one-to-one correspondence between the writing systems do not exist, and we report results based on our HTR experiments with two OT periodicals from the early twentieth century. Finally, we reflect on potential domain bias of HTR models in historical languages exhibiting spatio-temporal variance as well as the significance of working between writing systems for language communities that have experienced language reform and script change.

**INTRODUCTION**

Imagine that you are an Ottomanist starting a Ph.D. project or a new book. In addition to identifying a topic, doing a literature review, and fine-tuning your research question, you will also need to navigate an extensive archive of millions of documents that are not fully accessible as a repository. Large parts of it are yet to be cataloged, some collections have been opened to researchers only in recent decades, and digitization is at its earliest stages. To add to these challenges, the archive is written in a language whose script has been long abandoned and for which no digital tools, not even a keyword search option, are available. In these circumstances, you would want to create a scholarly apparatus for doing your work in a time-efficient manner that would allow you to comb through the most relevant sections of this archive. Where would you turn to annotate, or to transcribe for recall, these documents that were produced in a language that is old and in a script in which no one today, let alone the fledgling researcher, is perfectly literate?

In Istanbul, where the central archive of the Ottoman State and several of the most important research libraries in the field of Ottoman studies are located, there is an informal market of hireable transcribers to meet this demand. Consisting mainly of students and graduates of departments of history and Turkish literature, the informal labor of this market offers its services to scholars who do not have the time required to transcribe their documents. In the absence of extensive digitization of archives, they also provide on-site research services for their individual clients, many of whom do not live, or work regularly, in Istanbul.[1] Access to historical documentation in Ottoman Studies is, therefore, not unlike research in other countries and language situations decades ago.[2]

# Part 1 - Background & Historical Context

As the digitization of Ottoman-era archives is at present underway[3], our article concerns one approach to increasing computational access to them. We discuss Handwritten Text Recognition (HTR) applications for the automated transcription of Ottoman Turkish (OT), the Arabic-script language of administration and high culture in the Ottoman Empire between the fourteenth and early twentieth centuries.[4] While Turkish syntax and vocabulary formed the main framework of the language, OT borrowed extensively from Arabic and Farsi. With the political upheaval that ensued the demise of the Ottoman state and the creation of the Turkish Republic in the wake of WWI, OT was outlawed for public use in 1928 and supplanted by a "Turkified"[5] version of the language written from left to right (LTR) in Latin script [Lewis 2002].

This is not to say that the Ottoman archive was sealed in 1928. Physical access to it was possible, although limited [Aktaş 1992], but within a generation or so, the Arabic-script literacy required to use the archives declined. The first generation of historians who had been educated in Republican schools no longer had the linguistic skills of native speakers of OT to carry out research in the archives. Coupled with the ideological decision to shun the Islamic imperial heritage of the newly created nation-state of Turkey, the so-called Turkish language reform created deep gaps in the scholarly infrastructure of the early Republican period, with most serious academic discussions of the problems taking place outside of Turkey.[6]

Alphabet change had deeper cultural implications that went well beyond its ramifications for the scholarly establishment. Today, any Turkish citizen who wants to access their pre-1928 cultural heritage needs either to receive specialized education in Ottoman Turkish or to read the transcribed and edited versions of the select OT texts available in modern Turkish. There is, therefore, in modern day Turkey a real demand for transcription.[7] Furthermore, the language reform has not only created a barrier between one nation and its historical, cultural documentation but also, since OT was the former administrative language of the successor states of the Ottoman Empire, both language and script persist as barriers in accessing this larger imperial heritage by the broader scholarly community.

As interest in digital humanities turns to multilingual DH, the accessibility of archives and modes of participatory text creation, the Ottoman Turkish example is a thought-provoking case study because it links problems which the GLAM sector has traditionally confronted with problems not typically found in European language archives: text directionality and script, in particular.

## Archives and Access

For many languages of the world that have not traditionally been represented in the textual digital humanities, in particular historical languages, the obstacles of being "at the table" are very real.[8] Research materials and infrastructure for these languages are absent or in outdated formats, digitization is either partial or piecemeal, and accessing those digital resources can be a daunting task.[9] In the larger case of Ottoman Turkish, there is a lack not only of sufficient machine-readable corpora but also of natural language processing (NLP) tools designed to work with the multi-layered, multi-epoch complexities of this historical language.[10] When the archive is so abundant, yet still so inaccessible, the eagerness to open up research fields to exploration with digital methods grows every year.[11]

In this article, we argue that pursuing digital methods in some non-Latin script research fields, such as Ottoman studies, is not so much a question of resistance to, or distrust of, the methods as it is of the need for workable starting points, given the historical and linguistic complexity of these languages.[12] Furthermore, the desire for high-quality text creation at scale can stand at odds with the capacity and affordances of the platforms at our disposal to do so. Deferring the question of what kind of corpus might be representative of late Ottoman society for now, and building on the decades-long efforts to create open digital collections for reuse and high order analysis, what has come to be known as a "collections as data" approach [Padilla et al. 2019], we turn to deep learning methods for automatic transcription of printed Ottoman periodicals with the [Transkribus platform](). Whereas traditional Ottomanist communities have relied on workarounds for research that depend on informal labor for gaining access to archival materials, our case study makes the argument for technical workarounds within developing humanities research infrastructure that can, and do, spark innovation in the interest of linguistic inclusion.

## New Modes of Text Creation

At the intersection of archives and the textual digital humanities is the central question of access. Initiatives such as the EEBO-Text Creation Partnership have opened exciting windows for a wide range of scholarly projects grounded in access to historical print sources [Shawn 2007]. Its presence can be felt all over the scholarship in pre-modern English, and yet its mode of creation--outsourced double typing of texts in Asian countries--casts an uncomfortable, but often overlooked, shadow over the material conditions of its creation. Scholarly communities working on languages printed in the Arabic script (Arabic, Persian, Urdu, etc.) have been slower to arrive at robust textual digital humanities for multiple reasons, not the least of which has been the difficulties of converting the cursive-like printing of these languages into a machine-readable format [Ghorbaninejad et al. 2022]. We anticipate that examples in crowd-transcription, such as Transcribe Bentham (UCL), or more recent initiatives spearheaded within the GLAM sector (Library of Congress, Europeana, the Newberry Library, the Getty Museum), and the democratized platforms for engaging a larger public in doing so, will slowly make their way to cultural heritage institutions in RTL (right to left) language societies. What can be done in situations, however, such as OT, when the script of the archive is no longer native for the majority of the public? Our challenge has been not only that of automatic transcription but also rendering the output of this process sufficiently legible for a LTR (left to right) Turkish reader today.

Whereas some formal modeling has been done with OT (take for example, the markup of multilingual versions of the Ottoman Constitution in TEI XML including OT and Arabic versions [Grallert 2019]), our intervention focuses at a more rudimentary stage of new text creation; it is ground research that aims to produce someone of the first, if not the first, searchable text corpus of OT. This situation might be surprising to digital humanists working in European, or even large non-Western languages, for whom the existence of some kind of corpus has been a given for many decades. Unfortunately, Arabic-script OT missed the digitization moment that other languages enjoyed during the twentieth century.

One key technological development that opened a promising pathway for the creation of full-text searchable corpora for both printed and handwritten documents in Arabic script is the recent advances in pattern recognition with HTR. Unlike OCR, which operates at the level of individual characters, HTR works at line level and, does, therefore yield higher accuracy rates in recognizing languages such as Arabic.[13] In this article, we report on experiments we have carried out with HTR on periodicals printed in OT. The larger question of the full Ottoman archive, most of which is manuscript, however, still sits in the background of this article. Turn of the century printed materials in Ottoman Turkish (or in Arabic or Urdu, for matter), do exist on a continuum with handwritten materials, due to the fact that they are printed in a "cursive"-like manner, by which we mean that some of the letters are connected to each other. It is too simplistic, therefore, to oppose OT print and manuscript cultures, especially because Ottoman master printers did strive to reproduce the calligraphic traditions of the manuscript culture in the printed page [Özkal 2018, 71].

Our choice of applying HTR technology to OT print media is an intentional one, with the purpose of creating a workable access point to Ottoman historical materials that are already digitized. We worked with the automated transcription of two periodical series from the Hakkı Tarık Us (HTU) Digital Repository which holds the Ottoman Turkish periodicals published between the 1840s and the 1920s. With some 1,500 periodicals and an estimated number of 400,000 pages, the HTU online archive is perhaps the most comprehensive digital collection of the Ottoman Turkish press.The long term goal of our research is to generate general HTR models for text creation from this periodicals collection that will facilitate its use in computational research.

## Why Periodicals

After decades of digitization efforts in libraries and archives worldwide, the number of periodicals that are available to today's researchers has increased dramatically. For the historian, the serial periodical offers special insight into cultural debates as they unfold throughout time. Nineteenth-century periodicals are the place of remarkable emergence of public discourse in many regions of the world, including the multi-ethnic and multi-religious Ottoman Empire.[14] They also provide a valuable opportunity to reconsider this cultural sphere in a period marked by rapidly evolving linguistic usage in the face of political change [Mussell 2012]. Significant advances have been made globally in the accessibility of cultural collections through digitization and the implementation of full-text searchability and vibrant debates have opened up in digital history on how to use this digitized archive to the fullest extent.

There are a number of features of historical periodicals that make them complex for digital analysis in any language: OCR quality, the commercial nature of digitization collections, complex page layouts, the presence of images within the text, and particularities of their digital remediation [Ryan 2016; Nicholson 2013; Clifford et al., 2019]. Yet again, the challenge of dealing with Arabic-script periodicals is an even greater one. Whereas many periodicals have been scanned and brought together in digital archives in these languages, allowing researchers to access them at a distance, the texts of such periodicals are still not full text searchable, limiting users to slow reading and isolating this archive from computer-assisted modes of analysis.[15] Unsurprisingly, the few studies that have used various Ottoman periodicals collections as their main primary source have so far been qualitative and of limited scope. [Baykal 2019; Çakır 1994] The sheer volume of the available material, as well as the outdated formats and piecemeal nature of their digitization, renders distant reading of these sources unfeasible. These collections are not yet data.

# Part 2 - Transcription and Its Discontents

## Audience & A Pragmatic Approach

Mastering OT today is a demanding task. People who undertake learning the language/script nowadays do so in order to read historical manuscripts and documents. Among those who are proficient in OT, it is not common practice to take notes in Arabic script, but rather in what is the contemporary script habitus of Turkish historians as well as that of the majority of non-Turkish Ottomanists: the LTR Latin alphabet [Ghorbaninejad et al. 2022]; researchers in the field simply tend to annotate their documents in Latin script. On account of script difference and language change, annotation style can vary from person to person, with some researchers even transcribing their documents fully into modern Turkish. Furthermore, in Turkey, editions of OT works are rarely published in Arabic script, or if they are, they are accompanied by a facing-page transcription. As such, the Arabic script in contemporary Turkey could be considered a "dead script." Being able to automate OT transcription to Latin script, therefore, is a question of creating a knowledge infrastructure that corresponds to a practical issue of contemporary script literacy.

The usage of terms for discussing the passages between languages and scripts, from OT to MT, vary. Our Appendix (Key Concepts) details how we use the terms transliteration, transcription and romanization. Suffice it to say that the passage between writing systems--in history as well as in our use of HTR technology-- is far from a neutral one. We have come to realize that our approach to text creation needs to be a pragmatic one that balances a number of concerns. We have to pay enough attention to the requirements of the HTR engine so that the model is effective, keeping in mind the needs of the human reader, but also to be consistent enough so that the digital reuse of the resultant data is the highest quality possible. We are aware that, due to the particularities of OT, which we discuss in the following section about the encoding of grammatical information, it will probably never be possible to build an automated transcription system that can generate texts that are perfectly readable to the contemporary Turkish reader or that flawlessly bridge the gap between the original OT documents and their transcriptions. Our goal in applying neural models of automated transcription to OT printed materials is, instead, the creation of a "decent" text, a workable, "good-enough" approach to transcription, a goal echoed by other historians working in digitally under-resourced languages [Rabus 2020].

No doubt this will be disappointing to our philologically inclined colleagues, but it is the scale and the modes of downstream analysis that encourage us to adopt this position. We do not see the text creation process as a failure from the outset. Instead, working with automatic transcription has encouraged us to think about the transcription process itself as an initial access point to the historical materials. Let us be clear: our goal is to produce a result that is sufficiently accurate and usable by our target reader (the historian scholar) , and we recognize that the end result will never look like human-transcribed texts. From this starting point, it follows that what is necessary to use HTR methods to bring new kinds of access to a larger number of OT texts is a critical conversation about what is commonly called transcription.

We are particularly drawn to the idea of crowd transcription that could transform the existing informal economy of OT transcribers into a participatory community of shareable, open knowledge production. This will not be possible, however, without a larger scholarly debate about transcription and transliteration that departs from contemporary practices of inconsistent romanization in favor of one that is more harmonized with the ways that algorithmic systems work. Pilot studies such as ours with automatic transcription technologies are in a position to launch this debate. The question resembles one that is asked regularly in human-computer interaction research: how do we design interfaces and input mechanisms for human knowledge so that we can optimize the results of computational processes? The answers to such questions can only emerge, we believe, through the development of a user community that tests, evaluates and validates such transcription norms in different textual domains.

## Challenges of Transcribing OT and Its Implications for HTR Processing

Ottoman Turkish, in its classical form, is a patchwork of Turkish, Arabic, and Farsi vocabulary and grammar. The complexity of the language is compounded by the ambiguity of the Arabic script and the challenges that accompany its romanization [Halpern 2007]. Elezar Birnbaum summarizes the complex character of OT orthography as such :

"... the Ottoman Turkish writing system is only an *indication* of pronunciation rather than a representation of it. It incorporates two quite different methods of indicating sounds, which are ill-joined into one system… On the one hand, Arabic and Persian spelling conventions are preserved almost intact for all Ottoman words derived from those languages, while completely different conventions, rarely explicitly formulated and still more rarely consistently applied, even in a single piece of writing, hold the field for words of Turkish and other origins, and for Turkish affixes to Arabic and Persian loan words." [Birnbaum 1967, 123]

In contrast to the complicated nature of OT orthography, where several sounds in the language are omitted in writing and the correct pronunciation of a word is contingent upon the reader's literacy and linguistic background, Modern Turkish (MT) spelling is unequivocal and highly phonetic. This crucial difference between the two writing systems renders a one-to-one, diplomatic transliteration scheme from OT to MT unattainable.[16] This situation also partially accounts for the fact that to this day there is no scholarly consensus on how to, if at all, transcribe Ottoman Turkish to modern Turkish script.[17] However, if we remember the anecdote with which we began this article about the informal labor market of transcribers, despite the lack of a transcription scheme based on consensus[18], transcribed text is often preferred over original material as a matter of practicality.

To further break down the factors that hamper character-accurate, diplomatic transcription from OT script to MT:

1) There are only three vowel signs in Arabic ( ا, و, ى ) , which are often not represented but only implied in writing. MT, on the other hand, has eight vowels and the written word is always fully vocalized. Moreover, letters ( ا, و, ى ) are polyphonic, i.e., they correspond to more than one sound in MT script. For example, Arabic letter (و) may correspond to any of the following characters in MT : ( v, o , ö, u , ü ).

2) Several of the OT consonants are polyphonic as well. Depending on the textual context, they may be substituted by several different letters in MT alphabet.

| Ottoman Turkish | Modern Turkish |
|---|---|
| ا | a, e |
| ض | d, z |
| ك | k, g, ğ, n |
| و | v, o, u, ö, |
| ه | h, e, a |
| ى | y, a, ı, i |

Fig 1. Character correspondence chart of polyphonic OT letters

3) Finally, the phonological developments of Turkish language over the centuries and the wide variation in its pronunciation across the vast geography of the Ottoman Empire as well as across its socio-economic and ethnic groups at any given point in time, created a gap between its written and spoken forms. This means that many OT words may be pronounced in and transcribed to modern Turkish in several different forms, all of which might be considered accurate. Any endeavor to produce a "correct" transcription of OT in MT script is, in reality, an attempt to recreate what we imagine was the pronunciation of the Istanbul elite at the time, which, of course, is lost to us in its spoken form [Anhegger 1988].

For the purposes of our HTR experiments with OT, these three points have important practical implications. As an Ottoman Turkish text is transcribed into Latin script, linguistic information is added to its romanized version by way of vocalizing it and rendering it in one of its several alternative pronunciations, which might not necessarily correspond to the written form. The presence of polyphonic letters necessitates another layer of choices to be made by the transcriber. It is not unreasonable to regard the reading and transcribing of OT as an art rather than a science, a quality that does match the exigencies of the current generation of computational tools, including HTR. The unavoidable addition of linguistic information to the transcribed material is especially problematic when training a neural-network based transcription system such as *Transkribus*, where one-to-one correspondence between the original text and its transcription is essential. It also results in the introduction of a significant degree of bias to the language data that is provided by the transcriber for the neural network. We will discuss these issues in more detail in Part 3.

## Previous Work on the Automated Transcription of OT

There are currently two other projects that focus on the automated transcription of Ottoman Turkish [Korkut 2019; Ergişi et al., 2017]. Both are similar in methodology and scope, employing morphological analysis and lexicon-based approaches to romanize OT script. Korkut and Ergiş et al.'s studies take advantage of the fact that Turkish is an agglutinative language, the stems of Turkish words do not inflect when combined with affixes. Once an OT word is stripped of its affixes through morphological parsing, the stem can be looked up in a dictionary for possible matches in modern Turkish.[19] At the next step, these two romanization models reconstruct the stripped affixes in modern Turkish according to its spelling and pronunciation rules and offer potential transcriptions for the original OT word in Latin script. Both projects work with nineteenth and early twentieth century OT vocabulary, presumably because dictionaries with relatively extensive content are more readily available for this period than for previous centuries.

An important distinction of these romanization schemes from the HTR in *Transkribus* is their rule-based approach to automated transcription. Their systems depend on conventional linguistic knowledge (dictionaries and grammatical rules) whereas the HTR offers a statistical, brute force technique that utilizes large data sets for pattern recognition without any explicit reliance on linguistic information. While rule-based approaches might conceivably produce a higher precision output, Korkut and Ergişi et al.'s methods still have important shortcomings. In the absence of a rigorous OCR system for OT, it is not feasible to transcribe longer texts in their interfaces. They can also only romanize those words that are already in their databases. The scalability of these platforms depends on the creation of comprehensive OT to MT word-databases/dictionaries, which need to be significantly more substantial in content than anything that is currently available[20], and the development of effective OCR systems for the Arabic script to convert OT texts to machine-readable format. Even if these two conditions are met, however, the automated transcription of large corpora in these platforms might still be prohibitively slow due to the time needed for dictionary searches to match the OT stems to their MT counterparts.[21]

## HTR for Automated Transcription

In our study we use HTR in the *Transkribus* platform for the automated transcription of printed OT texts. In principle, an HTR engine can be taught to recognize any writing style from any language. To start, the system needs to be provided with training data to identify the specific patterns in a given body of text. In order to produce a reliable transcription model for the rest of the corpus, it is imperative to create an accurate "ground truth", a literal and exact transcription of a small body of sample text, that is representative of the corpus in question.

While the failure to take linguistic rules[22] into consideration in HTR training does create certain shortcomings in the quality of the transcription, we prefer the approach for its practicality, time-efficiency, and scalability. HTR is most effective with large corpora and after the initial investment of time for creating the ground truth, it is possible to automatically transcribe hundreds of pages within a matter of hours. The system is also flexible and accepts changes in parameters. That is, with a large enough training set, it is possible to generate general HTR models that will work for corpora produced with different handwriting styles or typefaces.[23]

A major advantage of using HTR for OT is its high accuracy rate at segmenting and recognizing cursive and connected writing styles. Unlike most OCR based systems, which operate at letter level (for both segmentation and recognition) and are, therefore, not very efficient for connected scripts such as Arabic, HTR works at line level and recognizes characters in the context of words and lines rather than as individual units. This renders it an excellent tool for languages written with the Arabic alphabet.

# Part 3 - Our Experiment

## The Procedure

The *Transkribus* platform produces automated transcriptions of text collections by using customized HTR models trained on partial transcriptions of the corpus. As we have argued at the beginning of Part 2, tools are often not a perfect fit for the historical source material we possess and the results we obtain teach us a significant amount about our objects of study. In this study, after significant trial and error, our research workflow was ultimately reverse engineered to address critical questions of both the object of study (the periodicals and their language) and the method itself.

We created two sets of training data and corresponding HTR models for two periodicals[24] from the HTU online collection, the above-mentioned digitized collection of OT periodicals. The first one of these publications is the *Ahali* newspaper, printed in Bulgaria between 1906 and 1907. For this publication, we adhered to a "loose transcription"[25] scheme that only indicates long OT vowels and does not use any other diacritical marks. The second periodical for which we generated a training set is *Küçük Mecmua*, published by the leading ideologue of Turkism at the time, Ziya Gökalp, between 1922 and 1923 in Diyarbakir, Turkey. This publication has a significantly "Turkified" vocabulary and a relatively standardized orthography compared to other OT publications. We applied the *Islam Ansiklopedisi* (IA) transcription scheme[26] for *Küçük Mecmua*, which uses diacritical marks to differentiate between polyphonic characters in the OT and MT alphabets. Our assumption was that the regularized orthography and the detailed transcription system might help to reduce ambiguity in the transcription text for this publication, providing us with a "cleaner" starting point for the HTR.

The neural network employed in the *Transkribus* platform "learns" to recognize a script by matching the characters, strings or words in the image files with their counterparts in the transcription. Character-accurate ground truth is, therefore, a prerequisite for creating reliable HTR models. This basic principle complicates the workflow for OT in *Transkribus* significantly for two reasons; first, as we mentioned above, the lack of one-to-one mapping between OT and MT scripts, and second, the opposite directionality of Ottoman Turkish documents and their Latin script transcriptions.

To address the first issue and minimize the inconsistencies in the training data, in our transcriptions we prioritized character accuracy over both the correct pronunciation of the OT words and some of the grammatical rules of modern Turkish.[27] We also opted for maintaining spelling mistakes and typos in the transcription text exactly as they appeared in the original pages, rather than correcting them. In other words, although these decisions seem counterintuitive to the experienced transcriber, working with an HTR system for a RTL historical language required us to unlearn some of the conventions of our "practical" research habits for the purposes of transcription for creating training data. By restraining ourselves and giving the machine what it needs to do its task, we hoped to optimize its performance.

The opposing directionalities of OT and MT come into play at the next step of the workflow, in which images of the documents are linked to corresponding transcriptions. While *Transkribus* does support RTL languages, it does not allow connecting RTL images to LTR transcription text. To avoid this problem, we devised a workaround and reversed the direction of our transcription text in modern Turkish from LTR to RTL.[28] To this end, we wrote a short script in Python and ran it for the plain text files of our transcriptions.

Following the upload of the image files, segmentation was performed automatically with the CITLab layout analysis tool in *Transkribus* for the recognition of text regions and baselines in the documents.  We then imported corresponding reversed-transcription texts into the platform.

Fig 2. A snapshot from *Transkribus* interface demonstrating the transcription process. Note the left-justified, yet reversed, Latin-alphabet transcription (center bottom) of the OT text (top right). The OT text displayed in the canvas tool is from *Küçük Mecmua.*

After the careful preparation of the two sets of ground truth with the workflow described above, sixty pages each, we generated our first HTR models for *Ahali* and *Küçük Mecmua*, yielding 9.84% and 9.69% Character Error Rates (CER) respectively. Whereas there was no significant difference in the CER between the two collections, in spite of the more detailed transcription scheme for the latter, in the second phase of the process, we retrained the models by manually correcting errors in the automated transcription and expanding each set of ground truth by an additional twenty pages of transcription after which the CER for *Ahali* newspaper dropped to 4.01% and for *Küçük Mecmua* this number was reduced to 3.63%. As it is clear from these numbers, the IA transcription scheme did not provide a significant advantage over the less detailed transcription system in terms of CER. This might, however, be due to the relatively modest size of the training data and a larger sample set might offer a better comparison amongst the various transcription models.

Finally, for text analysis purposes or human reading, the automatically transcribed texts are exported from the platform and their directionality is reversed to LTR.

## Cross-Domain Applicability

The *Tranksribus* platform allows the repurposing of HTR models to re-train the neural network for recognizing different corpora as well as community sharing of those models. In future work, we will take advantage of this function to develop general HTR model(s) and evaluate to what extent they work for the entire digital repository of the Hakkı Tarık Us (HTU) library. To this end, we will focus on creating additional training sets from the holdings of this digital collection.

For the HTR technology we have at our disposal, the HTU collection of late Ottoman periodicals is an ideal corpus. By the late 1870s the typeface for printing had been standardized and a set of printing conventions for OT periodicals established.[29] Furthermore, the OT press from this period tended to cover similar topics and promote comparable agendas; as a result, they contain similar vocabulary, terminology, and named entities. All of these factors, along with the possibility of using a "language model" function[30], contribute to the reduction of error rate in text recognition. It is this uniformity of this printed corpus that we hope will allow the HTR model to generalize across the entire HTU corpus with acceptable accuracy. It should be said though that for scholars in the field who intend to expand HTR to manuscript archives, finding a comparable corpus uniformity will no doubt be a challenge.

To test how the HTR model would function across different textual domains, we ran the *Ahali* HTR model for three other periodicals from the HTU collection. Attention was paid to include in the sample pool publications from different time periods with distinct content and agendas. The CER for random pages from these periodicals are as follows:

| Name of publication | Subject | Date of publication | CER |
| --- | --- | --- | --- |
| *Tasvir-i Efkar* | politics | 1863 | 8.54% |
| *Mekteb* | education and literature | 1891 | 10.26% |
| *Kadınlar Dünyası* | feminism | 1914 | 6.22% |

It is important to note here that the CERs listed above were attained without any training data from these publications or any correction and retraining of the model. This is an encouraging indication suggesting that with the expansion of our ground truth, it is, in fact, achievable to create viable general HTR model(s) for the entire online collection of the HTU library.

## Discussion of the Results

The primary challenge HTR faces when working with OT seems to be not character recognition, but rather identifying OT words in their proper Modern Turkish (MT) forms. As discussed earlier, the absence of one-to-one correspondence between OT and MT alphabets leads to multiple possible transcription outputs. Even when the HTR correctly identifies the characters in an OT word, accurate reading still depends on precise vocalization as well as context-appropriate pronunciation of the polyphonic letters. These, in turn, necessitate prior linguistic information, which is not taken into account during HTR training. In other words, no deep learning tool would be able to complete the desired task of a flawless MT transcription.

The recently implemented Language Models (LM)[31] function in *Transkribus* appears to partially compensate for the absence of rule-based text recognition in the platform. When supported with LM, HTR renders Ottoman Turkish words in their context-accurate form as they are defined in the training data. This, in turn, reduces Character and Word Error Rates.[32] For example, the OT word (عمله), which appears in our training set for the *Kucuk Mecmua* periodical, has two possible readings, both of which are character-accurate(imle) and (amele) whereas only the latter is a meaningful word unit in OT, the equivalent of the English word "worker". In our experiment, text recognition without Language Model transcribed the word as (imle). When we ran the HTR with the LM, however, the word was correctly recognized as (amele), the version that was defined in the training data. In addition, the LM appears to be able to identify character sequences that are not in the training data and even "assign a high probability to an inflected form without ever seeing it." [Strauss 2018, 6] This contributes to the accurate transcription of words that are not in the training set. While this is far from being the perfect solution for working around the ambiguities of OT, and a significant departure from rule-based NLP systems, it is still a step towards HTR-created, human-readable text .

Moreover, the neural network appears to be able to easily detect oft-repeated words and word pairs (such as Farsi possessive constructions, which are particularly challenging to identify in OT texts), presumably as a result of line-level operations of HTR. Consequently, it is reasonable to expect that the system will produce higher accuracy rates for document collections with formulaic language or corpora that revolve around similar subjects and, hence, use similar vocabulary. It is tempting to imagine a future of not only "language-blind" HTR, but also one that is language agnostic, or even multilingual adaptive, that can learn the particular linguistic strata or multilingual usage in a given corpus. It is still important to underline, however, that a more comprehensive solution to the problem of multiple possible outputs in OT transcription might be the integration of dictionaries/word indices into the system.[33] In the absence of such an option, the best alternative is to provide the HTR and LM with larger language data to account for a greater degree of orthographical variance.

Stemming from our experience with generating HTR models for different publications, we infer that the best approach to extensive collections, such as Hakkı Tarık Us, is to create date/period specific HTR models for subsets of those corpora. The late nineteenth century was a time of accelerated linguistic development and experimentation for Ottoman Turkish, both in its spoken and written forms. Therefore, temporal (and/or genre) proximity of publications in sub-groups is likely to contribute to the creation of more accurate HTR models. This would, presumably, also be the most practical approach to the automated transcription of the Ottoman manuscript archive, which is considerably more sizable than the print documents and publications in this language.

Finally, our experiment in the *Transkribus* platform affirmed the widely accepted conceptualization of transcription as a biased process and that machine facilitated transcription is no exception to this. Both HTR and LM rely on the language data provided by the transcriber; they, in turn, reproduce the transcriber's linguistic bias in the output. As aptly described by Alpert-Abrams:

"The machine-recognition of printed characters is a historically charged event, in which the system and its data conspire to embed cultural biases in the output, or to affix them as supplementary information hidden behind the screen." [Alpert-Abrams 2016]

In the case of the automated transcription of OT, the imposition of a uniform transcription scheme on the language obscures regional, temporal, and ethnic varieties in its pronunciations and creates an artificially homogeneous outlook for OT which does not reflect the historical reality of the language. For our project with OT periodicals, however, this point is less of a concern because these publications tended to, indeed, adhere to the standards of "high Ottoman", that is, to an OT spoken and written by the educated elites of the Empire. Still, the language information we, as the users of the automated transcription platform, impose on the neural network does not only affect the quality and readability of the output but also has important downstream implications from keyword searches and Named Entity tagging to other NLP applications to OT corpora. Corpora creation has for a long time been viewed as a complex political process, and with deep learning, computationally intensive models, this is no less the case.

# Conclusion

In our paper we have discussed ongoing research into text creation via automatic transcription in order to usher the larger OT archive into dialogue with analytical modes of the textual digital humanities. The limited development and application of OCR methods compatible with Arabic script languages has no doubt been one of the rate determining steps, not only in the development of textual corpora, but also in the attendant language technologies that support the use of such corpora. To wish--in 2020--for basic keyword search capacity within digitized media of the nineteenth and early twentieth century might seem to some as a rather old fashioned request, and yet for Ottomanists, it is an actual one. Limited access to digitized versions of the archive, a lack of language- and script-specific OCR in addition to a lack of scholarly infrastructure in the OT situation have meant that scholars have been slow to adopt digital methods.

The landscape of Arabic script languages is changing rapidly, especially with neural automatic transcription systems, as they can accommodate cursive-like scripts and typefaces. In order for these systems to perform well, however, they will require training on many more domains and spatio-temporal variants of the language and handwriting styles. In the case of printed materials, training on different typefaces and printing conventions will also be necessary.

A move to new transcription methods for Ottoman Turkish will not mean the total automatization of the transcription process, removing it from scholarly labor; instead it will require a reorganization of those efforts from informal to somewhat organized one. We do not expect that people will stop transcribing small amounts by hand for specific purposes, but we do suspect that it will change the way that large text creation projects work, as well as how archives create finding aids as digitized material becomes available.

We see the automated transcription of OT with *Transkribus* as a promising undertaking that could reach its full potential only by the participation of the larger community of Ottomanists. Currently, a major disincentive that holds back scholars in the field from investing their time in the process is the necessity to reverse the direction of the transcription text before and after processing it with HTR. As necessity can be the mother of invention, our "hack" to the system and our initial results with OT prompted the *Transkribus* development team to develop and integrate a new functionality into the interface that will automate reversing of the direction of the transcription text. It is our hope that this will support and encourage the wide-spread use of this technology for OT by the broadest number of researchers, as well as beyond OT to other multi-directional projects.

In the specific case of printed periodicals in OT, automatic transcription has long term promise. In order to achieve the goal of enabling keyword searching and more complex forms of modeling and analysis of OT textual content, a multi-pronged approach will be required. First, a team-based approach to transcription for the purposes of the creation of ground truth will be required (or even some crowdsourced, community-based approaches that have yet to be defined). Such a community effort might be achieved by sharing the model generated for the HTU print collection publicly, although how the endeavor would scale remains to be seen. Second, even though the platform-as-service model of the READ COOP may provide scholars with a variety of public models, the computing resources that allow these to run might exclude scholars and archives without the resources to buy in. It may be that only open models provide a more sustainable solution. Third, research is needed not only to recognize the Arabic script that is used in the specific genre discussed in this article--periodicals--but also, as in many other language situations, for structure recognition and image extraction that would allow for formal modeling of the resultant periodicals as research objects.

We are not experts in the histories of all writing systems, but our results may be generalized to other language situations that have changed, or have used multiple, scripts over time: Aljamiado, Azeri, Bosnian, Crimean, Judeo-Spanish, Malay and Wolof, just to name a few. Our research with automatic transcription of periodicals printed in the Ottoman Empire during the mid nineteenth to early twentieth century has important implications for multi-script archives of nation-states located in the borderlands of empires and religions. After all, language reforms forged new modernized versions of languages such as Hebrew and Arabic while maintaining their original scripts, but in other places writing systems were changed altogether. As we described in depth above, in Turkey the Arabic script of OT was abandoned in favor of a modified Latin alphabet. The political disintegration of states at other moments of the twentieth century, in particular the former Yugoslavia and Soviet Union, led to fragmentation of similar languages into script-divergent varieties that both straddle current political borders and co-exist with older variance found in archives. If script change comes about in moments of radical political change enacting ruptures in archives, it also seems to map onto a slow globalizing of the Latin script--and not without significant political debate--in different countries around the world around the turn of the twenty-first century. Turkmenistan adopted a Latin alphabet in the late 1990s as well as Kazakhstan as recently as 2017. This geopolitical context raises the question of what will become of digitized archives in such locales--printed and handwritten--and adds totally new contours to the notion of the "great unread" in nations already cut off from the scripts of their past, as well as those that may be facing a phasing out of old alphabets. Future developments in digitization and text creation from borderland archives--if they are to happen--need to take into account not only the automatic transcription of written language, but also the various writing systems in which those languages will be expressed.

## NOTES

[1] Such was the experience of one of the authors of this article, a native speaker of Turkish, while conducting her doctoral research on the emergence of a modern public education network for women in the Ottoman Empire. The examination of the undigitized and uncataloged documents of the Ottoman Ministry of Public Education and the relevant Ottoman Turkish periodicals took over a year. She enlisted the help of a transcriber to annotate and transcribe the documents she collected. The hired transcriber also conducted research in the Hakkı Tarık Us Periodicals Collection, which, at the time, had not yet been fully digitized.

[2] In a 1960 article, Stanford Show, a leading scholar of Ottoman history, described the challenges of working in Ottoman archives as follows: "Months of searching into the catalogs is necessary to locate all available materials concerning each subject, and much longer time is required to gather these materials and mold them into an intelligible unit of study" [Shaw 1960, 12]. For a more recent evaluation of the state-of-the-field, which reveals that there has not been any significant improvement in the accessibility of the archive since the 1960s, see [Gratien et al. 2014].

[3] The two most notable projects in the vein are the digitizations of the Hakkı Tarık Us (HTU) print periodicals repository and the manuscript collections of the Presidential Ottoman State Archives. HTU was fully digitized between 2003 and 2010 thanks to a collaboration between the Bayezid Library in Istanbul, the host institution to the HTU repository, and the Tokyo University of Foreign Studies. While the digitization of this vast collection is a momentous stride towards increasing accessibility of some of the key primary sources in the field of late Ottoman studies, its execution leaves much to be desired. The images of the periodicals were digitized as DjVu files, an outdated format that requires various workarounds to open on current personal computing platforms. Furthermore, even though it has been a decade since the completion of the digitization phase of the project, the web interface is still in Beta version. Catalogue information about the contents of the collection is meager and search options are non-existent. For a detailed report on the contents and accessibility of the HTU digital repository, see [Baykal 2011]. Note that, as of 2020, no improvements have been made in the interface to address the issues Baykal raises in his article. Work on the digitization of the central Ottoman archive continues to this day. So far, some 40 million documents from the collections of the Ottoman and Republican archives are estimated to have been digitized and made available online.

[4] As the small Ottoman principality from western Anatolia evolved into a multi-lingual, multi-ethnic empire that came to rule over most of the Middle East and South Eastern Europe the syntax and the

writing system of OT grew in their complexity while its vocabulary expanded exponentially. For an evaluation of the development of Ottoman language in its written form, see [Darling 2012].

[5] By "Turkified" we mean that many of the Arabic and Persian words in the language were removed in favor of words of Turkish origin. For an in depth analysis of Turkish language reform, see [Lewis 2002]

[6] As a case in point, two of the most prominent Turkish Ottomanists of the early Republican era, Kemal Karpat and Halil Inacik, received their higher education and/or worked in Western universities. A sizable portion of both of these scholars' publications are in English, i.e. their intended audience was Anglophone and not necessarily Turkish. For a further example of a key academic discussion taking place outside of Turkey, see Birnbaum's 1967 article on the transliteration of OT, where he discusses in detail the indifferent attitude in Turkey toward developing a coherent transcription scheme for OT [Birnbaum 1967].

[7] To this end, a team of professional transcribers is employed by the Presidential State Archives in Istanbul to transcribe, edit, and publish selections of documents from their vast holdings. These are made available online for free for public use. Turkish publishing houses also commission experts in the field to transcribe and annotate important cultural texts in Ottoman Turkish or to produce adaptations of these in modern Turkish for popular consumption.

[8] The conversation is not a new one, but recent years have seen growth in scholarly conversation about them. Researchers have recently stressed not only the urgency of access to digital tools but also the inequities of digital material and knowledge infrastructures across languages as a question of access. Notable efforts include a full-day workshop at the 2019 ADHO conference entitled "Towards Multilingualism In Digital Humanities: Achievements, Failures And Good Practices In DH Projects With Non-latin Scripts" (organized by Martin Lee and Cosima Wagner); the Multilingual DH Group and the efforts of Quinn Dombrowski; the RTL (Right to Left) workshop at the Digital Humanities Summer Institute organized by Kasra Ghorbaninejad and David Wrisley; and one-and-a-half day workshop "Disrupting Digital Monolingualism" organized by the Language Acts and Worldmaking Project.

[9] To this question of infrastructure, one really needs to add contemporary forms of inaccessibility that researchers--local and global--experience in times of reduced funding and mobility.

[10] While there is a small community of Turkish developers working on NLP tools for modern Turkish in open source environments, these are still far from being fully workable platforms. Neither are they sufficiently comprehensive to accommodate the complicated orthography of Ottoman Turkish. To our knowledge, the most advanced NLP tool for Turkish so far developed is Zenberek. Also see, [Sezer et al.] and [Sak et al. 2008].

[11] There have been several initiatives in recent years to introduce digital tools and methods to the field of Ottoman studies. The most notable is OpenOttoman, an online platform to foster collaborative DH scholarship among Ottomanists. For an overview of the objectives of the platform, see [Singer 2016]. The Baki Project and the Ottoman Text Archive Project are the two other noteworthy projects for creating textual DH platforms for Ottoman studies.

[12] An excellent argument for devising computational tools for such one such language, nineteenth century Hebrew, is [Soffer et al. 2020]. For a pilot study on the automated text recognition of nineteenth century printed books in Bengali, see [Derrick 2019]. Although not in a non-Western language, Ventresque et al.'s work on transcribing Foucault's reading notes with Handwritten Text Recognition

technology is another important case study that seeks a workable starting point for linguistically complex archival material [Ventresque et al. 2019].

[13] For an OCR system that is based on the same "line-level" recognition principle as HTR, see [Romanov 2017].

[14] See for example, [Baykal 2019], [ Karakaya-Stump 2003], [Gülaçar 2018].

[15] This is no less true of OT periodicals collection at HTU and Arabic periodicals in large collections such as [Arabic Collections Online (ACO)](#) and the [Qatar Digital Library (QDL)](#).

[16] Birnbaum discusses this issue in detail in his paper [Birnbaum 1967].

[17] In the context of debates surrounding a canonical seventeenth century OT text, H.E. Boeschton declared that "a fully consistent transcription is impossible" and "a transcription is a medium ill-suited for the presentation of linguistic results" [Boeschoten 1988, 25-26.] The position we take in this study, which is arguably the commonly accepted practice today, is formulated by Robert Anhegger as a response to Boeschton: "The guiding principle should be to produce a transcription which is as easy as possible to follow" [Anhegger 1988,14].

[18] Currently, there are four transcriptions schemes widely used in scholarly publications: Deutsche Morgenländische Gesellschaft (DMG) system, Encyclopaedia of Islam (EI, 2nd edition) system, International Journal of MIddle Eastern Studies (IJMES) system, and and Islam Ansiklopedisi (IA) system. For a comparative chart of these transcription systems, see [Bugday 2014]. Turkish scholars tend to prefer the IA system. Modern Turkish versions of OT texts that are produced for the general population, on the other hand, often do not adhere to a strict system but rather follow a "loose" transcription. As a case in point, the **Turkish Historical Society**- a government agency for the study and promotion of Turkish history - endorses a "loose" rather than "scientific" transcription of post-1830s printed works. For archival material from the same period, however, they recommend "scientific transcription", a term they use interchangeably with "transliteration". Türk Tarih Kurumu. "Eser Yayın Süreç: Çeviriyazı Metinlerde Uyulacak Esaslar." Accessed March 13, 2020. [https://www.ttk.gov.tr/eser-surec/eser-yayin-surec/](https://www.ttk.gov.tr/eser-surec/eser-yayin-surec/).

[19] The Arabic loanwords in OT complicates this scheme significantly as Arabic stems do inflect. Therefore, a comprehensive dictionary for such a system needs to include all inflected forms of the Arabic loanwords in OT.

[20] Korkut's system currently has only about 43.000 words while *Dervaze* boasts 72.400 OT words in its database

[21] Korkut points out a few long term solutions to this problem in his article [Korkut 2019, 5].

[22] Transkribus does allow integrating custom prepared dictionaries into HTR training, but at present the computational cost is too high to justify the miniscule improvement in the accuracy rate of the automated transcription. Based on a correspondence with the Transkribus team, however, we understand that there has been a recent improvement in the recognition-with-dictionary workflow which speeds up the process. In expectation of further developments in this vein, we have started compiling a comprehensive digital Ottoman Turkish dictionary based on Ferit Devellioğlu's seminal work [Devellioğlu 1998]. Once the dictionary is ready, it will be publicly shared on *Transkribus* platform.

[23] As a case in point, the number of general HTR models made available to the public in the Transkribus interface have increased significantly over the last year, expanding not only the language, but the domain and genre applicability of the method. For example, The National Archives of the Netherlands has published a composite model for Dutch-language documents of the seventeenth, eighteenth and nineteenth centuries called Ijsberg, trained on dozens of handwriting styles in commercial letters and notarial deeds, that has achieved a CRE rate of 5.15% [Transkribus 2019].

[24] We take these two publications to be typical OT periodicals in terms of page layout, typeface, and content. Transcriptions for the training data were repurposed partially from the research material of Suphan Kirmizialtin's dissertation study and partially from the website Osmanlıca Mahalli Gazeteler. Existing transcription texts were modified to normalize spellings, correct transcription errors, and standardize the transcription schemes.

[25] See Appendix.

[26] See note #18.

[27] This most frequently occured in cases that involve the conventionalized affixes of OT which violate the vowel harmony of MT. In such cases, while character-accurate transcription results in archaic-sounding pronunciation of those words (such as الدى : 'oldı'- 'happened' ; اوچونجى : 'üçünci''- 'third') they are still easily recognizable by modern readers; therefore, in those occasions, we opted to go with diplomatic transcription. However, in other instances, where literal transcription, i.e. transliteration, produces a pronunciation that renders the word unrecognizable to the speakers of the Turkish language (for example, خواجه : 'havace' ; كوكرجين : 'kükercin') we preferred the conventional pronunciation over character accuracy (خواجه : 'hoca' ; كوكرجين : 'güvercin'). This decision did, inevitably, introduce a certain degree of inconsistency to our HTR models.

[28] Here, we had to take into account the bi-directionality of OT. In OT, as in Arabic, alphabetical characters are written RTL while numerical characters are written LTR.

[29] By this time, *naskh* style typeface, cut by the revered punch cutter and master printer Ohannes Muhendisyan, had emerged as the standard continuous text typeface for OT publications. [Yazıcıgil 2015]

[30] For Language Models, see *Discussion of the Results* section.

[31] "Language Models (LM) estimate the probability of a specific word $w$ given the history of words $w_1$, $w_2$, … using external language resources. Assuming that these probabilities model the language of the current document well, we output the transcription which maximizes a combination of the HTR probability and the LM probability." [Strauss et al. 2018]

[32] For example, in our experiment with *Küçük Mecmua*, the CER was reduced from 5.23% to 3.63% when the automated transcription was implemented with the Language Model option.

[33] See note #22.

## APPENDIX- KEY CONCEPTS

*Transcription, Transliteration, Romanization*

There are a number of terms that are intrinsically related, and some of which we use interchangeably in this paper, that are central to our automated recognition of OT project.

*Transliteration* is the substitution of the characters of a script by those of another alphabet. In an ideal transliteration, each character of the source script will be represented with only one and always the same symbol in the target writing system, creating a one-to-one mapping, or graphemic correspondence, between scripts.

*Transcription*, on the other hand, is the representation of the sounds of an alphabet with the characters of another script. A transcription attempts to offer an accurate phonetic rendering of the source language even when this violates the graphemic correspondence between scripts. This means that, depending on the textual context, a polyphonic character- a letter that represents more than one sounds- may be substituted by several different letters in the target alphabet. In some cases, the reverse might be true. In the instance of OT transcription to modern Turkish, transcription also involves representing sounds, vowels to be more specific, which are only implied but are not written in the original text. Precisely because of this, the transliteration of OT inevitably becomes transcription.

**A *popular,* or "loose", *transcription***, is an approximation of the conventional orthography and popular pronunciation of a word in a different script. It will produce a transcription that is easy to follow by contemporary audiences while inevitably forgoing important linguistic information in the original text.

Finally, ***romanization*** is the substitution of the characters of a non-Roman script by those of the Roman alphabet [ Halpern, 2007].